\documentclass[journal]{IEEEtran}
\usepackage[utf8]{inputenc}
\usepackage{xcolor}
\definecolor{green}{rgb}{0,.8,0}
\usepackage{graphicx}
\usepackage{float}
\usepackage{cite}
\usepackage{slashbox}
\usepackage{amsmath,amssymb,amsfonts}
\usepackage{algorithmic}
\usepackage{makecell}

\ifCLASSINFOpdf
\else
\fi
\begin{document}
%
\title{Predicting the Timing of Camera Movements From the Kinematics of Instruments in Robotic-Assisted Surgery Using Artificial Neural Networks}

\author{Hanna Kossowsky,~\IEEEmembership{Student~Member,~IEEE}, and Ilana Nisky,~\IEEEmembership{Senior~Member,~IEEE}

\thanks{This study is supported by the Israeli Science Foundation (grant 327/20), the Israeli MOST (Israel-Italy virtual lab on “Artificial Somatosensation for Humans and Humanoids”), and by the Helmsley Charitable Trust through the ABC Robotics Initiative of Ben-Gurion University of Negev, Israel. HK is supported by the Lachish and Ariane de Rothschild fellowships.}
\thanks{Hanna Kossowsky and Ilana Nisky are from the Department of Biomedical Engineering and the Zlotowski Center for Neuroscience, Ben-Gurion University of the Negev, Beer-Sheva, Israel (e-mail: hannako@post.bgu.ac.il; nisky@bgu.ac.il).}}

\maketitle

\begin{abstract}
Robotic surgeries offer many benefits, however do not allow for simultaneous control of the endoscopic camera and the surgical instruments. This leads to frequent interruptions as surgeons adjust their viewpoints. Autonomous camera control could help overcome this challenge. We propose a predictive approach for anticipating when camera movements will occur using artificial neural networks. We used kinematic data of surgical instruments from robotic surgical training. We split the data into segments, and labeled if each segment immediately preceded a camera movement or did not. Due to the large class imbalance, we trained an ensemble of networks on balanced sub-sets of the data. We found that the instruments' kinematics can be used to predict when camera movements will occur, and evaluated the performance on different segment durations and ensemble sizes. We also studied how much in advance upcoming camera movements can be predicted, and found that predicting camera movements up to 0.5 s in advance led to only a small decrease in performance relative to predicting imminent camera movements. These results serve as a proof-of-concept for predicting the timing of camera movements in robotic surgeries and suggest that an autonomous camera controller for robotic surgeries may someday be feasible.
\end{abstract}


%
\IEEEpeerreviewmaketitle

\section{Introduction}
\IEEEPARstart{R}{obotic}-assisted minimally invasive surgeries (RAMIS) have gained popularity over the past decades \cite{hussain_use_2014, lanfranco_robotic_2004}. During RAMIS, e.g. with the da Vinci surgical system (Intuitive Surgical Inc., Sunnyvale, California), the surgeon uses Master Tool Manipulators (MTMs) to control a camera arm, and Patient Side Manipulators (PSMs), on which surgical instruments are mounted; the surgeon uses a foot pedal to switch between their control \cite{guthart2000intuitive}. RAMIS offer many advantages to both patients and surgeons, compared to open or laparoscopic surgeries. Patients benefit from less postoperative pain, shorter hospital stay, reduced complication rates and less tissue damage and blood loss \cite{hussain_use_2014, motkoski_toward_2013}. Surgeons benefit from reduced tremors compared to open surgeries \cite{lanfranco_robotic_2004}, as well as more degrees of freedom, better visualization, and improved ergonomics compared to laparoscopic surgeries \cite{kenngott_robotic_2008, lanfranco_robotic_2004}. 

Despite the many benefits offered by RAMIS, they are not without disadvantages, 
such as an increase in the operative time and the need for special training \cite{hussain_use_2014}. 
Furthermore, unlike open surgeries, similar to all minimally invasive surgeries, in RAMIS the visual feedback is limited; only a specific region of the surgical scene is visible to the surgeon, who controls an endoscopic camera arm to adjust the viewpoint \cite{pandya_review_2014}. Studies have shown that the camera is moved frequently, especially by expert surgeons \cite{jarc2017viewpoint}, and the value of good camera control has been demonstrated in several works \cite{medina1997image, conrad2006role, way2003causes, jarc2017viewpoint}. Poor visualization and sub-optimal viewpoints, on the other hand, can have detrimental effects; they can lower surgeons’ awareness to the surgical environment and lead to an increase in surgical errors, causing injuries and impairing patients' safety \cite{gallagher2009ergonomic, pandya_review_2014, mariani_experimental_2020}. 
Simultaneously controlling the camera and the surgical instruments is not possible in RAMIS \cite{pandya_review_2014, dardona_remote_2019}, necessitating surgeons to release control of the instruments every time they wish to re-position the camera \cite{eslamian2020development, dardona_remote_2019, pandya_review_2014}. Such manual re-positioning can be time consuming \cite{hussain_use_2014, eslamian2020development}, lead to disruptions in the flow of the surgery \cite{eslamian_autonomous_nodate} and can potentially be distracting \cite{eslamian2016towards}.

A potential solution to this challenge is the automation of the camera movements \cite{wijsman_first_2018, da2020scan}. One approach for such automation is tracking methods. For example, tracking the movements of the surgical instruments and moving the camera such that the instruments are always in the field of view \cite{mariani_experimental_2020, eslamian_autonomous_nodate, eslamian2016towards, wijsman_first_2018, molnar2020visual, wijsman2021efficiency}, or in the surgeons' preferred area \cite{li2020data}. This can be beneficial, as it is important that the surgical instruments remain within the field of view to avoid injuries and tissue damage \cite{mariani_experimental_2020, eslamian2016towards, molnar2020visual}. 
While tracking the surgical instruments has been shown to be a feasible and beneficial method, it assumes the instruments to be the only deciding factor in the surgical scene, while in reality, additional features, such as anatomical features, likely affect the optimal viewpoint \cite{rivas2019transferring, ko2005intelligent}. 
Hence, the instruments could still be in the field of view, but not necessarily be what define the camera movements at every given moment. This is supported by \cite{ko2005intelligent}, in which they designed a robotic camera system for laparoscopic surgeries that could switch between two modes: track instrument, or show specific anatomical site based on the current surgical stage. 
Other camera automation tracking approaches are based on tracking the surgeon's eyes \cite{ali2008eye}, head \cite{dardona_remote_2019} or head and and upper-body motions \cite{gao2021augmented} to control the camera. There are several common characteristics between these methods. One is that they necessitate adding tracking systems to the autonomous camera controller. Additionally, in many of these methods, the camera is constantly moving to follow the instruments, eyes or head, whereas in RAMIS, the camera is often stationary and moved at specific times. To prevent the constant movement of the camera, it may be possible to define a threshold that needs to be surpassed for camera movements to occur. This would, however, necessitate devising a method for defining the threshold. Additionally, the tracking methods focus on reactive, rather than predictive, camera automation.

Several works have investigated the potential for predictive camera control. One such study positioned the camera in accordance with anticipated instrument trajectories that were predicted using Markov models \cite{weede_intelligent_2011}. Furthermore, several works have designed algorithms for predicting camera viewpoints and movements using data collected in dry-lab tasks \cite{agrawal2018automating, ji2018learning, rivas2019transferring}. For example, in \cite{agrawal2018automating}, participants performed a pick-and-place task while controlling simulated camera movements with their heads; inverse reinforcement learning was then used to predict the camera movements.
However, the dry-lab tasks in these studies \cite{agrawal2018automating, ji2018learning, rivas2019transferring} are more structured and less complex than surgical procedures. Wagner et al., \cite{wagner2021learning} designed an autonomous camera guiding robot for laparoscopic surgeries based on a full surgical procedure. Their work showed promising results, however, was performed using a human model, which too is different than a real anatomical setting. To the best of our knowledge, very few works have focused on the prediction of RAMIS camera movements in real and complex surgical environments.  

When approaching the subject of camera automation, the question of the control method arises. For example, while tracking the surgical instruments has been shown to improve different aspects of users' performance, it assumes that the camera control method should be to follow the instruments. However, it is currently unknown what exactly drives surgeons' camera movements and their timings. Therefore, an alternative method for automating camera movements in RAMIS can be to use the surgeon as the model to imitate using artificial neural networks.  
Artificial neural networks have been used in various works in RAMIS; for example, in the segmentation of surgical procedures into gestures \cite{itzkovich2019using, ramesh2021multi, wu2021cross} and the classification of surgeons' skill levels \cite{fawaz_accurate_2019}.
However, to the best of our knowledge, predicting camera movements from RAMIS data using artificial neural networks was not demonstrated to date.

Different forms of RAMIS data can be used to assess the possibility of predicting camera movements in RAMIS. One option is to use the video stream, whereas another is to use the kinematic data of the surgical instruments. 
Using the video stream offers several advantages, such as the comprehensiveness of the data, i.e., everything the surgeon sees is reflected in the video stream. Additionally, when using systems that involve moving an endoscopic camera, the video stream will always be available. 
On the other hand, using the kinematics of the instruments can allow for using smaller and simpler neural networks than those required by the video stream. This can be advantageous in RAMIS, as online running time can be increased due to an increase in the network complexity. 
Furthermore, if the instruments' kinematics can be used to predict camera movements in RAMIS, it can shed light on the relation between instruments and camera movements from the perspective of understanding surgeons’ sensorimotor control. 

We hypothesize that it may be possible to use the instruments' kinematics to predict the camera movements. Previous works have shown autonomous camera systems which track the movements of the instruments to be beneficial \cite{mariani_experimental_2020, eslamian_autonomous_nodate, eslamian2016towards, wijsman_first_2018}. While these benefits can be related to factors such as a reduction in the mental workload, it can also indicate a relation between the camera and the instruments movements. This relation is logical, as the surgeon performs the surgery with these instruments, and removing them from the field of view can lead to injuries \cite{mariani_experimental_2020, eslamian2016towards, molnar2020visual}. Furthermore, the kinematics of the instruments reflect the surgeons' hand movements. In open surgeries, the surgeons move their eyes and head to adjust their viewpoint, which may be reflected by the camera movements in RAMIS. Head, eye and hand movements have been shown to be related \cite{carnahan_temporal_1991}. 
Hence, we posit that the kinematics of the surgical instruments may be indicative of upcoming camera movements. 


The problem of predicting the camera movements in RAMIS can be divided into two sub-problems: (1) predicting the timing and (2) predicting the trajectories of the camera movements. In this study we use artificial neural networks to predict the timing of the camera movements from the kinematics of the surgical instruments, recorded during surgical training using the da Vinci surgical system. 
We approached the problem of predicting the timing of camera movements as a classification problem. We split the kinematic data of the instruments into segments, and labeled each segment either as one that immediately precedes a camera movement, or one that does not. We then trained neural networks to classify segments of kinematic data either as immediately preceding a camera movement or not immediately preceding a camera movement. 

A critical challenge when predicting the timing of the camera movements is that the data is highly imbalanced: there are many more kinematic segments that do not immediately precede a camera movement than those that do. Imbalanced data impairs the ability of neural networks to learn, as they tend to misclassify minority class samples as belonging to the majority class, due to the latter's increased prior probability \cite{johnson2019survey}. One solution is under-sampling the majority class, such that the network is trained on a balanced subset of the data. However, this solution discards much of the available data, and the subset of samples drawn from the majority class may not be representative. Another solution, used in sentiment classification \cite{zhang2015ensemble} and credit scoring \cite{wang2015large}, is to train an ensemble of networks, each on a balanced subset of the data, and to then combine the predictions of all the networks. Each of the networks receives all the minority class samples and a different subset of majority class samples, which equals in size to the number of samples in the minority class. 

In this study, we use an ensemble of neural networks to analyze kinematic data that was recorded during surgical training on porcine models with the da Vinci surgical system. Our results indicate that the timing of the camera movements can be predicted from the kinematic data. We examine the performance of the ensemble for several ensemble sizes and segment durations. Furthermore, in addition to the prediction of imminent camera movements, we examine the possibility of advance prediction of upcoming camera movements to open the possibility of online prediction of camera movement timing during RAMIS. Finally, we discuss directions for future improvements of our ensemble.

The contributions of our work are as follows:
\renewcommand{\labelitemi}{\tiny$\blacksquare$}
\begin{itemize}
  \item This work serves as a proof of concept for the possibility of predicting the timing of camera movements in RAMIS, and shows that this is possible using solely the kinematics of the surgical instruments, with no video input.
  \item We examine several kinematic segment durations and demonstrate which contains the most relevant information for the prediction of camera movements in RAMIS.
  \item We show that, in addition to predicting imminent camera movements, advance prediction is possible, indicating that online prediction of the timing of camera movements in RAMIS can potentially be possible. 
    \item We show that predicting camera movement events is possible in unstructured and complex data recorded during surgical training using the da Vinci surgical system. 
\end{itemize}

\section{METHODS}
\subsection{Data}
The data used in this work are recordings of surgical training on a porcine model using the da Vinci surgical system, provided to us through a collaboration agreement with Intuitive Surgical Inc. The protocol, titled "Computer Enhanced Minimally Invasive Surgery-Surgeon and Staff Training" was approved on July 31, 2019. The dataset consists of recorded procedures performed by seven surgeons (three experts and four novices), each performing two tasks, Uterine Horn Dissection and Simulated Cuff Closure. That is, a total of 14 recorded surgical tasks were used in this work. 

We used the kinematic data of the three surgical instruments that were used in these tasks. The instruments were mounted on the three PSMs (Patient Side Manipulators), the kinematics of which were sampled at 50Hz. As in \cite{dipietro2016recognizing, itzkovich2019using}, the kinematic data we used were the endpoint position in the three Cartesian axes, the three linear velocities, and the gripper angle of each PSM. There were therefore a total of 21 features per sample. The velocity was numerically calculated by differentiating the position according to the time. We then filtered the position and velocity data using second order zero-phase Butterworth filters, with cutoff frequencies of 5Hz and 8Hz, respectively. Finally, we normalized each of the features by subtracting its mean and dividing it by the standard deviation (z-score normalization). 

\subsubsection{Imminent Camera Movement Prediction}

\begin{figure}[!htb]
\centering
\includegraphics[width=\linewidth]{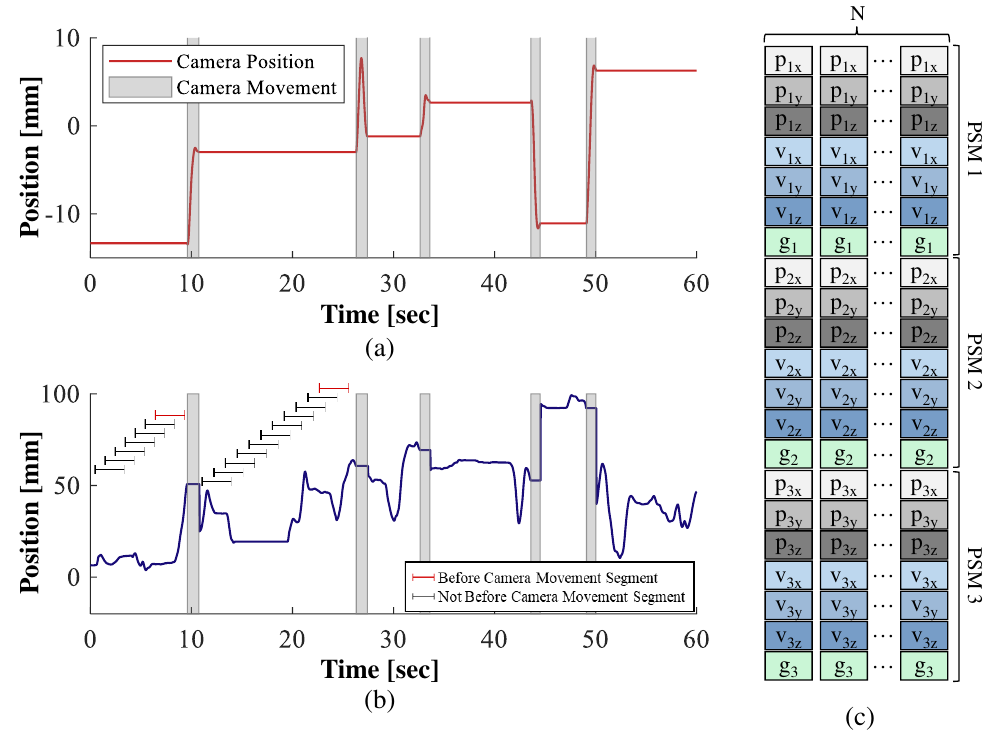}
\caption{Data Preparation. (a) The position of the camera in one of the axes. The gray shaded areas indicate the timestamps in which camera movements were detected. (b) The position of one of the PSMs in one of the axes relative to the camera position shown in (a). Note that the instrument position is recorded relative to the camera, leading to the different position values immediately before and after a camera movement. The splitting of the kinematic data into segments is illustrated, where segments that immediately precede a camera movement are indicated in red, and those that do not are black. Data from timestamps during which there was a camera movement were not included in the segments. (c) An illustration of a sample (network input). The colored rectangles represent the 21 kinematic features, seven for each of the three PSMs. The length of each feature is N samples. $p_{ij}$ represents the position of the i\textsuperscript{th} PSM in the j axis, where i = 1, 2, 3 and j = x, y, z. Similarly, $v_{ij}$ represents the velocity, and $g_{i}$, the gripper angle.}
\label{DataSetup}
\end{figure}

To label the data, we used the camera arm's endpoint Cartesian position to identify the timestamps in which the camera moved, and in which it was stationary. 
Fig \ref{DataSetup}(a) shows an example of the camera's endpoint position in one of the axes, where the timestamps identified as those in which the camera moved are highlighted in gray. 
We then split the kinematic data of the PSMs into segments, and labeled each segment either as immediately preceding a camera movement (designated a \textit{before camera movement segment}) or not immediately preceding a camera movement (designated a \textit{not before camera movement segment}). 
Hence, the inputs to the neural networks were kinematic data, comprised of 21 features, each with a length of N samples. 
Fig \ref{DataSetup}(b) shows an example of the position of one of the PSMs in one of the axes (i.e., one of the 21 kinematic features), 
with a loose illustration of the splitting of the kinematic data into segments. The overlap between every two consecutive segments was N-1.

We chose to split the kinematic data into segments rather than input the signals into the networks and train them to recognize the sample immediately preceding camera movements due to the large class imbalance. Had we inputted the signals, the networks would be trained on a great deal of samples not immediately preceding camera movements and very few that do. Splitting the data into segments allowed us to input any number of segments from each of the two classes into the neural networks.

We aimed to find the segment duration that would lead to the highest success in predicting the timing of camera movements on our dataset, and therefore created several versions of the segmented kinematic data, each with a different segment length, N = 25, 50, 100 and 200 samples, corresponding to segment durations of 0.5, 1, 2 and 4 seconds. Fig. \ref{DataSetup}(c) illustrates a segment (network input), comprised of 21 features, each of length N.

\begin{figure*}[!htb]
\centering
\includegraphics[width=17cm]{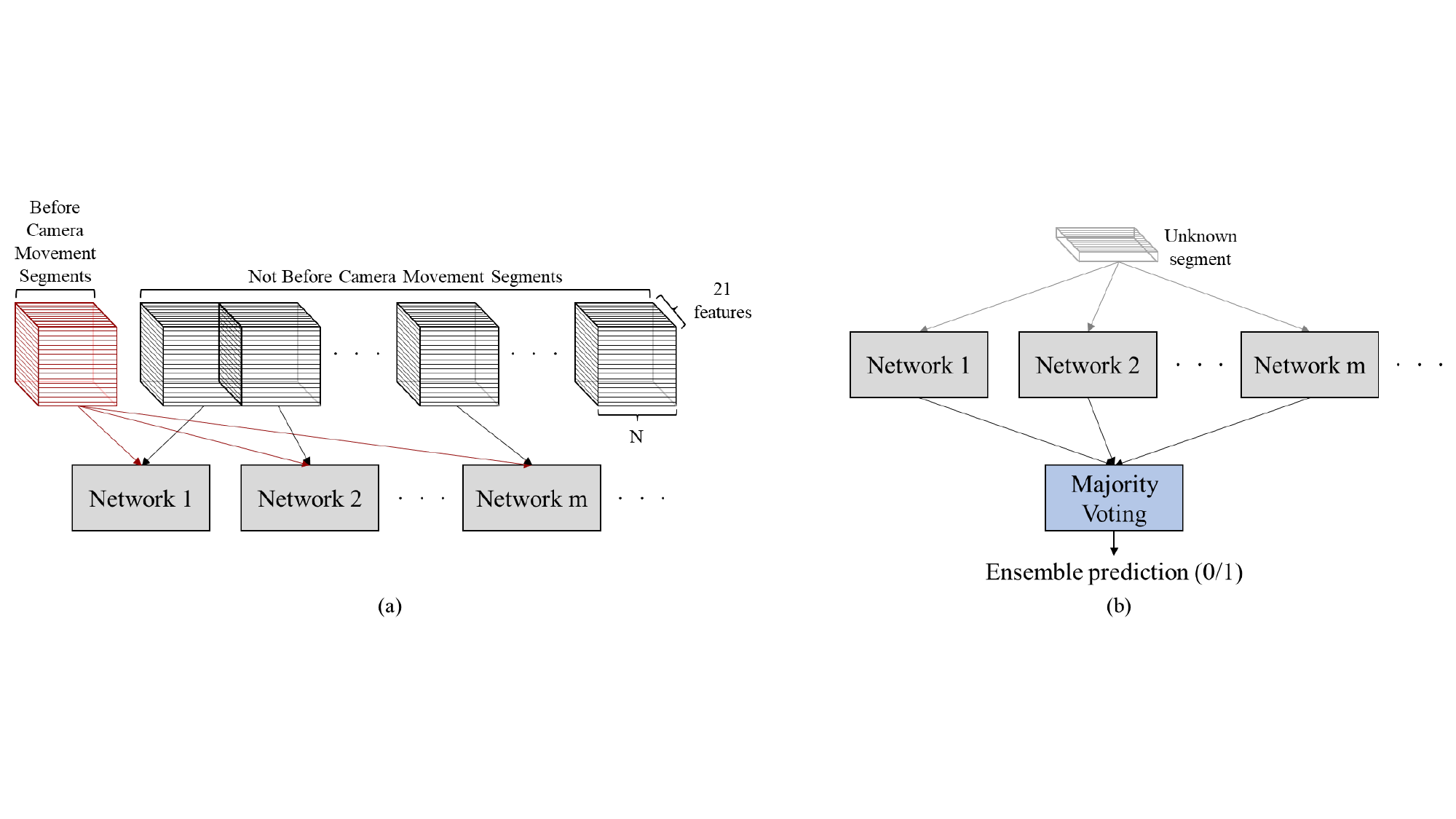}
\caption{Ensemble of networks. (a) Illustration of the training process. Each of the networks in the ensemble received all the \textit{before camera movement segments} in the training set and an equivalent number of \textit{not before camera movement segments} from the training set. (b) Illustration of the prediction stage. Each of the networks in the ensemble received the unknown segment and outputted its prediction (0 for a \textit{not before camera movement segment} or 1 for a \textit{before camera movement segment}). The prediction of the ensemble was determined using a majority vote of the predictions of all of the networks in the ensemble.}    
\label{Ensemble_explanation}
\end{figure*}

We created \textit{test}, \textit{validation} and \textit{train} sets for each of the four segment durations. The \textit{test} sets were created by randomly selecting 15\% of the \textit{before camera movement segments} from each recording (in the event of a not whole number of segments, we rounded up), and an equivalent number of \textit{not before camera movement segments}, thus achieving balanced \textit{test} sets. This process was repeated using the remaining data to create balanced \textit{validation} sets. The remaining, unbalanced data were our potential \textit{train} sets. All the \textit{before camera movement segments} in the \textit{train} set were used to train each of the networks in the ensemble, while a different subset of \textit{not before camera movement segments} was used for each network (Fig \ref{Ensemble_explanation}(a)). This allowed us to use a larger portion of the data, while training each of the networks on a balanced dataset. 

The datasets created for each of the segment durations were of different sizes (Table \ref{tab:datasetsizes}). For example, if two camera movements occurred less than 200 samples apart, this camera movement would not be included in the 4 s duration dataset, but may be included in the shorter duration datasets. We chose not to downsample all our datasets to the smallest size as our dataset was very small to begin with. We acknowledge that smaller datasets can lead to poorer performance; however, our main goal in this work was to assess the possibility of predicting the timing of the camera movements using the kinematics of the instruments. We therefore decided to use all the data available to us in order to answer the question, can the kinematics of instruments be used to predict when camera movements will occur, to the best of our ability. As we aimed to achieve the highest performance we could with our data, we aimed to find the segment duration that would lead to the best performance for our dataset, and therefore, examined the performance of different segment durations to find which would be best in this case.

\begin{table}[htb!]
  \begin{center}
    \caption{Dataset Sizes}
    \label{tab:datasetsizes}
    \begin{tabular}{|c|c|c|c|} 
      \hline
      \backslashbox{\textbf{Data set}}{\textbf{Segment Duration}} & \textbf{Test} & \textbf{Validation} & \textbf{Train}\\
      \hline
      \textbf{0.5 s} & 134 & 120 & 566\\
      \hline
      \textbf{1 s} & 130 & 112 & 534\\
      \hline
      \textbf{2 s} & 120 & 104 & 478\\
      \hline
      \textbf{4 s} & 106 & 94 & 420\\
      \hline
    \end{tabular}
  \end{center}
\end{table}

We additionally examined the potential for overlapping segments in our datasets. There were many more \textit{not before camera movement segments} than \textit{before camera movement segments}. Specifically, the ratio between the two was on the order of magnitude of $10^3$. Hence, due to the random sampling of the \textit{not before camera movement} segments, most of the segments did not overlap with one another. However, there was a larger percentage of segments with overlap for the longer segment duration datasets. 
Specifically, the percentage of overlapping segments for the 0.5 s segment duration was under 7\%, where only 3\% of them had an overlap of 50\% or higher. The overlap for the 1 s duration was under 13\%, where 6\% had an overlap of 50\% or higher. In the 2 s duration, the overlap was 22\%, with 12\% with an overlap of 50\% or higher. Lastly, the overlap for the 4 s duration was under 38\%, where under 23\% had an overlap of 50\% or higher.

As a subset of the \textit{not before camera movement segments} was used to train each of the networks in the ensemble, the segments were not fed into the networks in a time-consecutive order. Furthermore, to prevent overfitting and biases, random shuffling was used, such that the order of the segments was different in each training epoch. Hence, the networks were trained to receive a kinematic segment and recognize if it would, or would not, be immediately followed by a camera movement. Such a trained network could then potentially be used in RAMIS; the network could continuously receive data segments and indicate those that should be followed by a camera movement. 

\subsubsection{Advance Camera Movement Prediction}
We additionally examined the possibility of predicting that a camera movement would occur in a certain amount of time, rather than immediately following the kinematic segment. This is important for opening the possibility of online execution of an automated camera movement following the prediction that a camera movement should occur. We therefore examined the performance of the ensemble when trained to predict that a camera movement would occur in 0.25, 0.5 and 1 second, and compared these results to the ability of the ensemble to predict an imminent camera movement. 

Here, our main interest was to assess the cost in performance that would be caused by advance prediction. That is, if imminent prediction is possible, could this prediction method potentially be used online? To investigate this possibility, we examined the relation between the amount of advance prediction time and the decrease in performance. We therefore aimed to isolate the factor of advance prediction time and to ensure that our results would stem from that alone. This came across in the segments included in this part of our work. If two camera movements were too close together, there could arise a situation in which there would not be enough samples in between to allow for advance prediction. Therefore, in this part, we took only the \textit{before camera movement segments} which could belong to all the datasets. Hence, in this part, we used a smaller portion of our dataset, such that there were 478 segments in our training data. 

\subsubsection{Feature Contribution}
To examine the contribution of each of the features (position, velocity, and gripper angle) to the performance of the ensemble, we removed each in turn and trained each of the ensemble's networks on the remaining features. When either the position or the velocity was omitted, there were 12 features, and when the gripper angle was excluded, there were 18 features. The goal of this analysis was to shed light on the features that may have contributed more to the learning of the neural networks in this task.

\subsection{Networks}
\subsubsection{Network Evaluation}
To evaluate the ensembles' performances, we used three metrics: accuracy, True Positive Rate (TPR), and True Negative Rate (TNR). 
The accuracy is defined as the number of correctly classified segments divided by the total number of classified segments. 
\begin{equation}\label{accuracy}
    Accuracy = \frac{TN+TP}{TN+FP+FN+TP}
\end{equation}
where True Negative (TN) is the number of \textit{not before camera movement segments} correctly classified as such. Similarly, True Positive (TP) is the number of \textit{before camera movement segments} correctly classified as such. False Positive (FP) refers to the case where a \textit{not before camera movement segment} is predicted to be a \textit{before camera movement segment}, and False Negative (FN) is when a \textit{before camera movement segment} is predicted to be a \textit{not before camera movement segment}. 

The additional two metrics we examined reflect the ability of the ensemble to detect a \textit{before camera movement segment} when there is one (TPR), or a \textit{not before camera movement segment} when there is no imminent camera movement (TNR): 
\begin{equation}\label{TPR}
    TPR = \frac{TP}{TP+FN}
\end{equation}
\begin{equation}\label{TNR}
    TNR = \frac{TN}{TN+FP}
\end{equation}

We ran each of the networks in this work 10 times with different random seeds to ensure that the results did not vary greatly between the different seeds. 
To test the statistical significance of our results, we used ANOVA Models with the MATLAB statistic toolbox (2018a). Following that, we compared between every two conditions using post-hoc t-tests. We used the Holm-Bonferroni correction for multiple comparisons. We present the p-values after this correction ($p_{corrected}$), and therefore the threshold significance level following the correction is 0.05. 

\subsubsection{Ensemble Size and Segment Duration}
Each of the networks in the ensemble was an LSTM network (Long Short-Term Memory network) \cite{hochreiter1997long}, as this network architecture is able to learn sequential data due to a memory unit. 
To choose the ensemble size, based on \cite{itzkovich2019using} and several pilots, we began with a two LSTM layer network. We examined the performance of nine network architectures to choose the ensemble size: each LSTM layer could be comprised of either 100, 300 or 500 neurons. To stabilize the learning process and reduce over-fitting we added a dropout layer with a probability of 0.2 after each LSTM layer, and a batch normalization layer \cite{ioffe2015batch} between the two LSTM-dropout blocks. We used the adam optimizer with a learning rate of 0.0001, and decayed the learning rate with every epoch according to:
\begin{equation}\label{lr_decay}
    lr_i = decay*lr_{i-1}, 
\end{equation}
where $lr_i$ is the learning rate in epoch $i$, and $decay=0.99$. 
 
For small ensemble sizes, each additional network has a larger effect on the ensemble's performance; this effect decreases as the ensemble size increases. We tested several ensemble sizes, and the number of composing networks was chosen such that stable performance of the ensemble was achieved \cite{wang2015large}. To combine between the outputs of the networks, we used majority voting \cite{hernandez2013large}: each of the networks predicted the segment's class, and the ensemble's output was the class predicted by the majority of the networks (Fig \ref{Ensemble_explanation}(b)).

After choosing the ensemble size, we then assessed which of the segment durations would lead to the best performance of the ensemble. We continued with the previously described nine networks for this stage. We examined the accuracy, TPR and TNR of the ensembles for the four different segment durations, and selected the duration that produced the best results when predicting when camera movements would occur. 

\subsubsection{Hyperparameter Tuning}
After finding the ensemble size that consistently led to stable results for the nine network architectures, as well as the segment duration that consistently led to the highest performance for our data, we further tuned the hyperparameters of the network for this ensemble size and segment duration. This was to obtain the highest performance we could with our selected segment duration considering our small dataset.
We tested the use of one or two LSTM layers, the number of neurons in each layer (100, 300, 500, 700, 900 and 1100), dropout after each LSTM layer (0.1, 0.2 and 0.3), recurrent dropout in each LSTM layer (0.0 and 0.2), the number of batch normalization layers (zero, one and two), learning rate (0.001 and 0.0001), learning rate decay rate (0.90, 0.99 and 1), batch size (32, 64, 128 and 256), and L2 regularization (0.1, 0.01, 0.001 and 0.0001).

The weights of the LSTM layers were initialized according to the number of neurons in the layer, by selecting their values from a normal distribution, $N(0,\sqrt{\frac{2}{LayerSize}})$. We monitored the \textit{validation} set loss, and used early stopping to end the training if the loss did not decrease for three consecutive epochs. We additionally evaluated the performance of the ensemble when using L1 regularization and when normalizing the data according to its maximal value, but both these options yielded poorer results and we do not report them here. We also assessed the performance when averaging the outputs of the networks in the ensemble instead of majority voting, but this too led to poorer performance. 

\section{RESULTS}
We chose the ensemble size, segment duration and network hyperparameters by training the described networks on the \textit{train} set, and testing them on the \textit{validation} set. 
Once arriving at our final ensemble architecture, we tested its performance on the yet untouched \textit{test} set, and present these results at the end of this section. 
\subsection{Ensemble Size}
\begin{figure}[!htb]
\centering
\includegraphics[width=\linewidth]{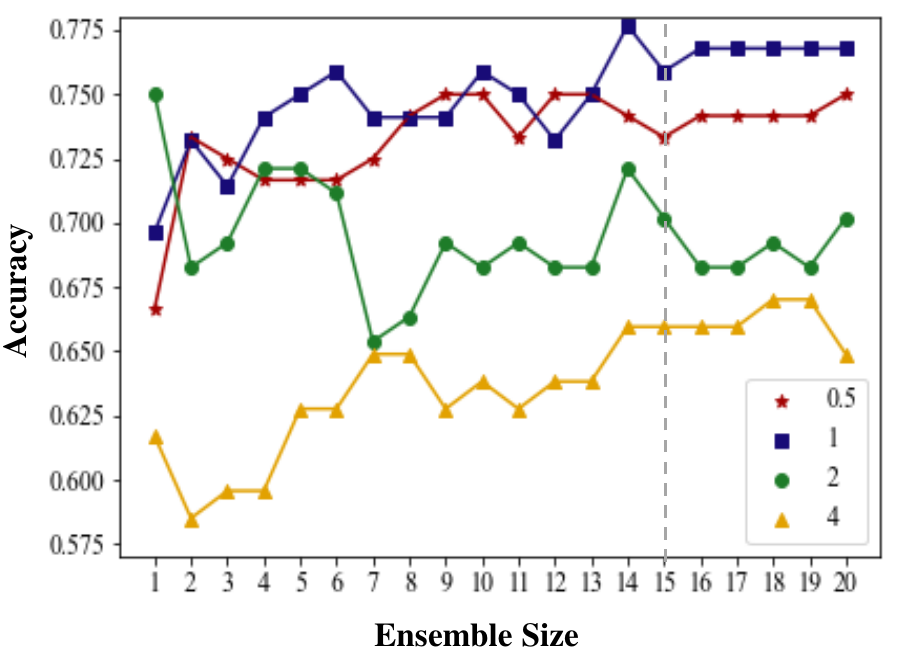}
\caption{Accuracy as a function of ensemble size. This graph shows the accuracy of the ensemble as a function of the number of networks in the ensemble. Each color and symbol represents the performance on one of the four segment durations: 0.5 s – red stars; 1 s – blue squares; 2 s – green circles; and 4 s – yellow triangles. The dashed gray line marks the ensemble size of 15, beyond which the addition of further networks had a relatively small effect on the performance.}    
\label{Ensemble_Size}
\end{figure}

We evaluated nine ensemble architectures to choose the ensemble size. Fig. \ref{Ensemble_Size} shows the accuracy of one ensemble as a function of the ensemble size, for each of the four segment durations. For smaller ensemble sizes, every added network affected the accuracy of the ensemble. As the ensemble size increased, this effect became smaller, such that for ensemble sizes of 15 and higher, the accuracy was relatively stable. This was observed for all nine ensembles, and similar stabilization was observed in the TPR and TNR values as well. We therefore selected an ensemble size of 15 networks.

\subsection{Segment Duration}
\begin{figure*}[!htb]
\centering
\includegraphics[width=16cm]{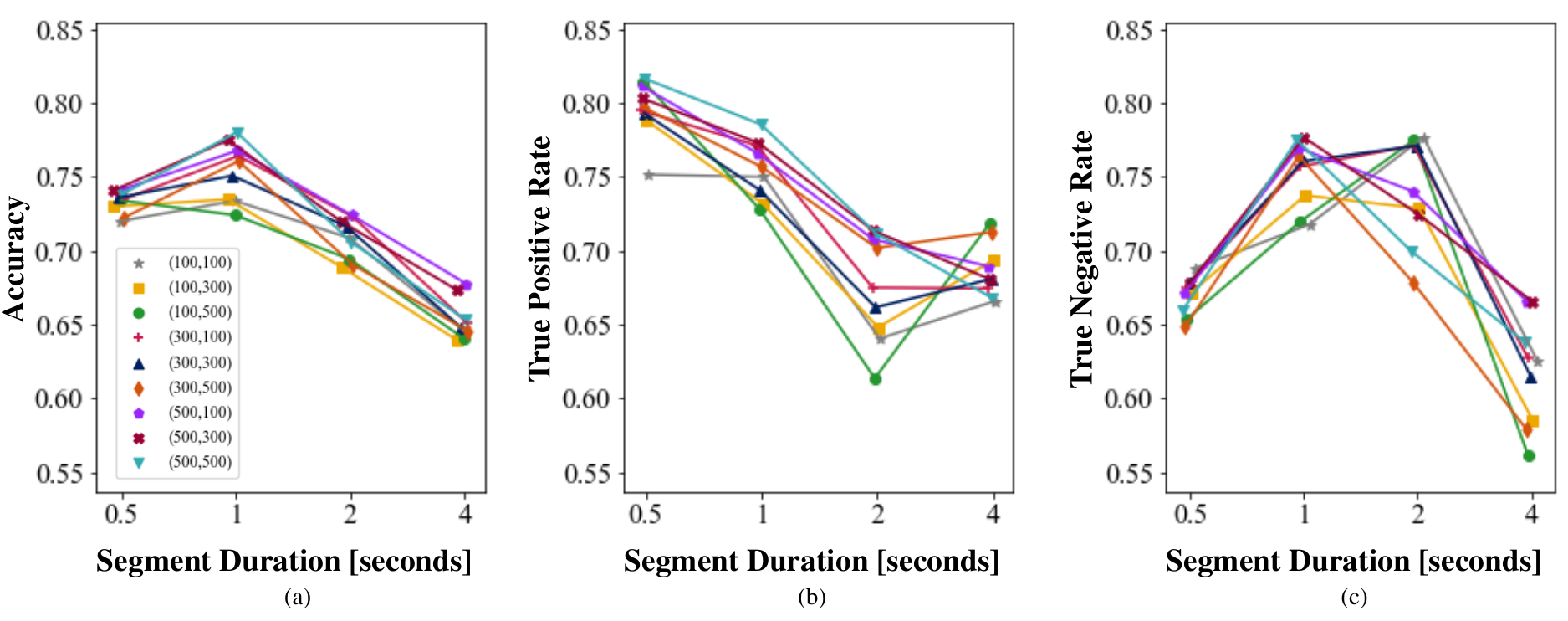}
\caption{Model performance as a function of the segment duration. The abscissa is the segment duration, and the ordinate is the average (a) accuracy, (b) True Positive Rate, and (c) True Negative Rate. Each color and symbol represents one of the nine different ensembles, which differ in the sizes of the two LSTM layers of the networks of which the ensemble is composed (as shown in the legend in (a)).}    
\label{Seg_length}
\end{figure*}

Fig. \ref{Seg_length} shows the average performance of the 10 evaluations of the nine ensembles, each of which was comprised of 15 networks, for each segment duration. All the standard deviations of the 10 repetitions were smaller than 0.1.
The accuracy achieved for the segment durations of 0.5 s and 1 s were higher than those of 2 s and 4 s (Fig. \ref{Seg_length}(a)). A comparison between the 0.5 s and 1 s segments revealed that eight of the nine ensembles scored higher for the 1 s segments. Next, Fig. \ref{Seg_length}(b) showed that the TPR values of the 0.5 s segments were higher than those of the 1 s segments. Similar to the accuracy, the TPR values for the 0.5 s and 1 s segments were higher than those of 2 s and 4 s.
When examining Fig. \ref{Seg_length}(c), we saw that the TNR values of the 1 s segments were higher than those of the 0.5 s. 
We also noted that there were cases in which the 2 s segments had the highest TNR, however as this duration was inferior in both the accuracy and the TPR, it was not chosen.

To choose between the 0.5 s and the 1 s segment durations, we took all three metrics into account. First, in most of the cases, the 1 s segments had higher accuracy values. Second, we aimed to design an ensemble that would be able to both detect when there would be, and when there would not be, a camera movement. Hence, both the TPR and TNR values were of equal importance. 
We saw that the TPR and TNR values for the 1 s segments were similar (Fig. \ref{Seg_length}(b-c)), whereas the 0.5 s segments had a high TPR, but a low TNR. We therefore selected the 1 s segment. We note that had we found either the shortest or longest segment duration to be the best of the four, we would have expanded our segment duration options, but this was not the case.

We additionally assessed the statistical significance of the effect of the segment duration on the accuracy, TPR and TNR using three nested ANOVA models. The factor of the 10 repetitions was nested in the factor of the network architecture. The results are summarized in Table \ref{tab:statistics_duration}. We observed a significant effect of the segment duration on all three metrics. Furthermore, the comparisons between every two durations were significant. That is, the 1 s segment duration led to significantly higher performance than the other three durations in terms of accuracy and TNR, and yielded higher TPR values than the 2 s and 4 s segment durations.

\begin{table}[htb!]
  \begin{center}
    \caption{Effect of Segment Duration on Performance}
    \label{tab:statistics_duration}
    \resizebox{\columnwidth}{!}{
    \begin{tabular}{|c|c|c|c|}
      \hline
      \backslashbox{\textbf{Test}}{\textbf{Metric}} & 
      \multicolumn{1}{|p{3cm}|}{\centering \textbf{Accuracy}} & \multicolumn{1}{|p{3cm}|}{\centering \textbf{TPR}} & \multicolumn{1}{|p{3cm}|}{\centering \textbf{TNR}}\\
      \hline
      \textbf{ANOVA model} & \thead{$F_{(3, 267)}=448.16$ \\ $p<10^{-4}$} & \thead{$F_{(3, 267)}=255.79$ \\ $p<10^{-4}$} & \thead{$F_{(3, 267)}=267.78$ \\ $p<10^{-4}$} \\
      \hline
      \textbf{0.5 s vs. 1 s} & \thead{$t_{(267)}=7.32$ \\ $p_{corrected}<10^{-4}$} & \thead{$t_{(267)}=7.96$ \\ $p_{corrected}<10^{-4}$} & \thead{$t_{(267)}=15.30$ \\ $p_{corrected}<10^{-4}$}\\
      \hline
      \textbf{0.5 s vs. 2 s} & \thead{$t_{(267)}=8.63$ \\ $p_{corrected}<10^{-4}$} & \thead{$t_{(267)}=23.88$ \\ $p_{corrected}<10^{-4}$} & \thead{$t_{(267)}=13.04$ \\ $p_{corrected}<10^{-4}$}\\
      \hline
       \textbf{0.5 s vs. 4 s} & \thead{$t_{(267)}=27.38$ \\ $p_{corrected}<10^{-4}$} & \thead{$t_{(267)}=21.45$ \\ $p_{corrected}<10^{-4}$} & \thead{$t_{(267)}=9.36$ \\ $p_{corrected}<10^{-4}$}\\
      \hline
      \textbf{1 s vs. 2 s} & \thead{$t_{(267)}=15.95$ \\ $p_{corrected}<10^{-4}$} & \thead{$t_{(267)}=15.92$ \\ $p_{corrected}<10^{-4}$} & \thead{$t_{(267)}=2.26$ \\ $p_{corrected}=0.025$}\\
      \hline
       \textbf{1 s vs. 4 s} & \thead{$t_{(267)}=34.71$ \\ $p_{corrected}<10^{-4}$} &  \thead{$t_{(267)}=13.48$ \\ $p_{corrected}<10^{-4}$} & \thead{$t_{(267)}=24.66$ \\ $p_{corrected}<10^{-4}$}\\
      \hline
      \textbf{2 s vs. 4 s} & \thead{$t_{(267)}=18.75$ \\ $p_{corrected}<10^{-4}$} &  \thead{$t_{(267)}=2.44$ \\ $p_{corrected}=0.015$} &  \thead{$t_{(267)}=22.40$ \\ $p_{corrected}<10^{-4}$}\\
      \hline
    \end{tabular}
    }
  \end{center}
\end{table}

\subsection{Hyperparameter Tuning}
We examined the accuracy, TPR and TNR of each of the networks architectures and arrived at the final hyperparameters. 
Our final network architecture was two LSTM layers, the first with 1100 neurons, and the second with 300. Each of these layers had a recurrent dropout of 0.0, and was followed by a dropout layer with a probability of 0.2. One batch normalization layer was placed between the two LSTM-dropout blocks. L2 regularization was used in both LSTM layers, with a value of 0.001. The learning rate was 0.0001, with a decay rate of 0.99, and the batch size was 128. 


After selecting the hyperparameters, we ensured that for this architecture too, the 1 s segment performed best. We therefore trained this architecture on each of the four segment durations, and compared between the results obtained on each of their validation sets. As shown in Table \ref{tab:FinalArchitectureComparison}, our final architecture showed the same trend as the previously checked nine. 

\begin{table}[h!]
  \begin{center}
    \caption{Final Architecture Comparison Between Segment Durations}
    \label{tab:FinalArchitectureComparison}
    \resizebox{\columnwidth}{!}{
    \begin{tabular}{|c|c|c|c|c|}
      \hline
      \backslashbox{\textbf{Metric}}{\textbf{Segment Duration [sec]}} & 
      \multicolumn{1}{|p{1.8cm}|}{\centering \textbf{0.5}} &
      \multicolumn{1}{|p{1.8cm}|}{\centering \textbf{1}} &\multicolumn{1}{|p{1.8cm}|}{\centering \textbf{2}} & \multicolumn{1}{|p{1.8cm}|}{\centering \textbf{4}}\\
      \hline
      \textbf{Accuracy} & ${0.74} \pm {0.02}$ & ${0.78} \pm {0.01}$ & ${0.73} \pm {0.01}$ & ${0.67} \pm {0.01}$\\
      \hline
      \textbf{TPR} &  ${0.81} \pm {0.02}$ & ${0.79} \pm {0.02}$ & ${0.73} \pm {0.03}$ & ${0.69} \pm {0.03}$\\
      \hline
      \textbf{TNR} & ${0.67} \pm {0.03}$ & ${0.78} \pm {0.03}$ & ${0.74} \pm {0.02}$ & ${0.65} \pm {0.02}$\\
      \hline
    \end{tabular}
    }
  \end{center}
\end{table}

\subsection{Imminent Camera Movement Prediction}
The average performance of the 10 runs of the ensemble comprised of these networks on the \textit{test} set was ${0.72} \pm {0.02}$ accuracy, a TPR of ${0.74} \pm {0.02}$ and a TNR of ${0.70} \pm {0.04}$ in predicting the timing of the endoscopic camera movements. The average confusion matrix of the 10 runs is presented in table  \ref{tab:TestConfusionMatrix}. This table shows the average and standard deviation number of segments classified as TN, FP, FN and TP.
\begin{table}[h!]
  \begin{center}
    \caption{Average Test Set Confusion Matrix}
    \label{tab:TestConfusionMatrix}
    \begin{tabular}{|c|c|c|}
      \hline
      \backslashbox{\textbf{True Class}}{\textbf{Predicted Class}} & 
      \multicolumn{1}{|p{1.4cm}|}{\centering \textbf{Not Before} \\\textbf{Camera}\\\textbf{Movement}} & \multicolumn{1}{|p{1.4cm}|}{\centering \textbf{Before} \\\textbf{Camera}\\\textbf{Movement}}\\
      \hline
      \textbf{Not Before Camera Movement} & ${45.3} \pm {2.4}$ & ${19.7} \pm {2.4}$\\
      \hline
      \textbf{Before Camera Movement} & ${17.3} \pm {1.5}$  & ${47.7} \pm {1.5}$ \\
      \hline
    \end{tabular}
  \end{center}
\end{table}\\

\begin{figure*}[!htb]
\centering
\includegraphics[width=16cm]{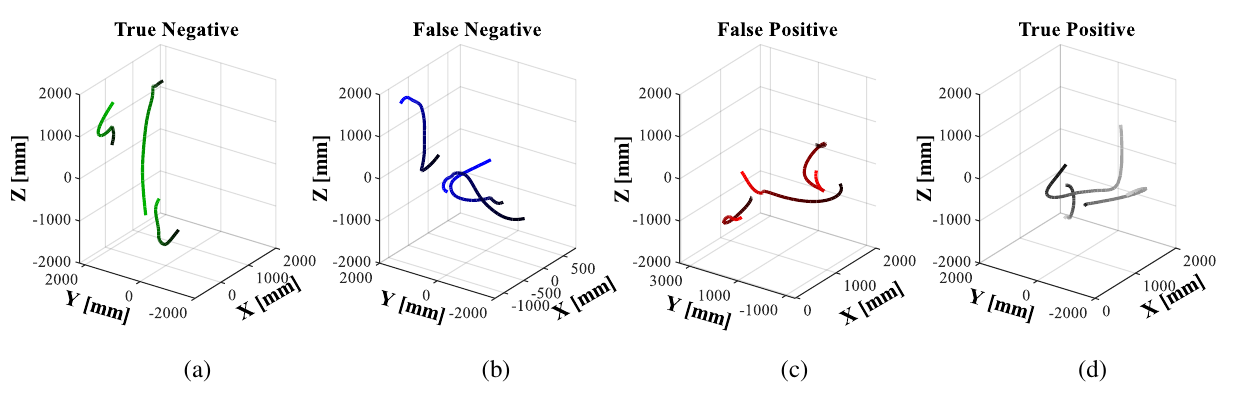}
\caption{Classification Examples. This figure shows examples of segments correctly and incorrectly classified as \textit{before camera movement segments} and \textit{not before camera movement segments}. The displayed signals are the position of one of the PSMs, recorded relative to the camera. Each plot shows three examples of segments, and the color indicates the direction of the movement, beginning with the light shades, and ending with the dark. (a) True Negative, (b) False Negative, (c) False Positive, and (d) True Positive.}    
\label{classification_results}
\end{figure*}

Fig. \ref{classification_results} presents examples of the classification of several segments from the \textit{test} set. Fig. \ref{classification_results}(a) shows three segments that were correctly classified as \textit{not before camera movement segments}, Fig. \ref{classification_results}(b) shows segments that were incorrectly classified as \textit{not before camera movement segments}, Fig. \ref{classification_results}(c) shows segments that were incorrectly classified as \textit{before camera movement segments}, and Fig. \ref{classification_results}(d) shows segments that were correctly classified as \textit{before camera movement segments}. We found no clear visual distinction between segments classified as \textit{before camera movement segments} and those classified as \textit{not before camera movement segments}. Specifically, the instruments were not idle before camera movements, and did not appear to exhibit any visible patterns of preparation towards the upcoming camera movements. 

\subsection{Advance Camera Movement Prediction}
We compared the ensemble's ability to predict an imminent camera movement with its ability to predict an upcoming camera movement in advance. The architecture of these ensembles was identical to that which predicted imminent camera movements, as we were interested in assessing the cost in performance caused by advance prediction. 
Table \ref{tab:AdvancePredictionAbsolute} shows the performance of the four ensembles. 
Table \ref{tab:AdvancePrediction} shows the ability of the ensemble to predict an upcoming camera movement 0.25, 0.5 and 1 second ahead of time relative to the imminent prediction achieved with the smaller dataset in terms of accuracy, TPR and TNR.

We used ANOVA models to examine the effect of the advance prediction time on the accuracy, TPR and TNR, using three separate analyses (Table \ref{tab:statistics_advance}). We found that advance prediction of 0.25 s did not significantly decrease the accuracy relative to imminent prediction, whereas advance prediction of 0.5 s and 1 s did. Furthermore, advance prediction of 0.25 s decreased the TPR, however increased the TNR, relative to imminent prediction. Predicting camera movements 0.5 s in advance significantly decreased the TPR, and advance prediction of 1 s decreased both the TPR and TNR values.

\begin{table}[h!]
  \begin{center}
    \caption{Advance Prediction Performance}
    \label{tab:AdvancePredictionAbsolute}
    \resizebox{\columnwidth}{!}{
    \begin{tabular}{|c|c|c|c|c|}
      \hline
      \backslashbox{\textbf{Metric}}{\textbf{Time [sec]}} & 
      \multicolumn{1}{|p{1.8cm}|}{\centering \textbf{0}} &
      \multicolumn{1}{|p{1.8cm}|}{\centering \textbf{0.25}} &\multicolumn{1}{|p{1.8cm}|}{\centering \textbf{0.5}} & \multicolumn{1}{|p{1.8cm}|}{\centering \textbf{1}}\\
      \hline
      \textbf{Accuracy} & ${0.674} \pm {0.010}$ & ${0.662} \pm {0.009}$ & ${0.632} \pm {0.019}$ & ${0.565} \pm {0.018}$\\
      \hline
      \textbf{TPR} &  ${0.723} \pm {0.013}$ & ${0.667} \pm {0.018}$ & ${0.647} \pm {0.016}$ & ${0.578} \pm {0.024}$\\
      \hline
      \textbf{TNR} & ${0.625} \pm {0.013}$ & ${0.657} \pm {0.011}$ & ${0.618} \pm {0.029}$ & ${0.552} \pm {0.019}$\\
      \hline
    \end{tabular}
    }
  \end{center}
\end{table}

\begin{table}[h!]
  \begin{center}
    \caption{Advance Prediction Performance in Percentages Relative to Imminent Prediction}
    \label{tab:AdvancePrediction}
     \resizebox{\columnwidth}{!}{
    \begin{tabular}{|c|c|c|c|}
      \hline
      \backslashbox{\textbf{Metric}}{\textbf{Time [sec]}} & 
      \multicolumn{1}{|p{1.2cm}|}{\centering \textbf{0.25}} & \multicolumn{1}{|p{1.2cm}|}{\centering \textbf{0.5}} & \multicolumn{1}{|p{1.2cm}|}{\centering \textbf{1}}\\
      \hline
      \textbf{Accuracy} & 98\% & 94\% & 84\%\\
      \hline
      \textbf{TPR} & 92\% & 89\% & 80\%\\
      \hline
      \textbf{TNR} & 105\% & 99\% & 88\%\\
      \hline
    \end{tabular}
    }
  \end{center}
\end{table}

\begin{table}[h!]
  \begin{center}
    \caption{Effect of Advance Prediction on Performance}
    \label{tab:statistics_advance}
    \resizebox{\columnwidth}{!}{
    \begin{tabular}{|c|c|c|c|}
      \hline
      \backslashbox{\textbf{Metric}}{\textbf{Test}} & 
      \multicolumn{1}{|p{3cm}|}{\centering \textbf{Accuracy}} & \multicolumn{1}{|p{3cm}|}{\centering \textbf{TPR}} & \multicolumn{1}{|p{3cm}|}{\centering \textbf{TNR}}\\
      \hline
      \textbf{ANOVA model} & \thead{$F_{(3, 27)}=89.58$ \\ $p<10^{-4}$} & \thead{$F_{(3, 27)}=101.29$ \\ $p<10^{-4}$} & \thead{$F_{(3, 27)}=39.86$ \\ $p<10^{-4}$} \\
      \hline
      \textbf{0 s vs. 0.25 s} & \thead{$t_{(27)}=1.71$ \\ $p_{corrected}=0.098$} & \thead{$t_{(27)}=6.74$ \\ $p_{corrected}<10^{-4}$} & \thead{$t_{(27)}=3.20$ \\ $p_{corrected}=0.007$}\\
      \hline
      \textbf{0 s vs. 0.5 s} & \thead{$t_{(27)}=5.72$ \\ $p_{corrected}<10^{-4}$} & \thead{$t_{(27)}=9.12$ \\ $p_{corrected}<10^{-4}$} & \thead{$t_{(27)}=0.67$ \\ $p_{corrected}=0.506$}\\
      \hline
       \textbf{0 s vs. 1 s} & \thead{$t_{(27)}=14.98$ \\ $p_{corrected}<10^{-4}$} & \thead{$t_{(27)}=17.24$ \\ $p_{corrected}<10^{-4}$} & \thead{$t_{(27)}=7.42$ \\ $p_{corrected}<10^{-4}$}\\
      \hline
      \textbf{0.25 s vs. 0.5 s} & \thead{$t_{(27)}=4.00$ \\ $p_{corrected}<10^{-3}$} & \thead{$t_{(27)}=2.38$ \\ $p_{corrected}=0.025$} & \thead{$t_{(27)}=3.88$ \\ $p_{corrected}=0.002$}\\
      \hline
       \textbf{0.25 s vs. 1 s} & \thead{$t_{(27)}=13.26$ \\ $p_{corrected}<10^{-4}$} &  \thead{$t_{(27)}=10.50$ \\ $p_{corrected}<10^{-4}$} & \thead{$t_{(27)}=10.62$ \\ $p_{corrected}<10^{-4}$}\\
      \hline
      \textbf{0.5 s vs. 1 s} & \thead{$t_{(27)}=9.26$ \\ $p_{corrected}<10^{-4}$} &  \thead{$t_{(27)}=8.13$ \\ $p_{corrected}<10^{-4}$} &  \thead{$t_{(27)}=6.75$ \\ $p_{corrected}<10^{-4}$}\\
      \hline
    \end{tabular}
    }
  \end{center}
\end{table}

\subsection{Feature Contribution}
We removed each of the features (position, velocity, and gripper angle) from the data in turn, and examined the performance of the ensemble. We used the 1 s segment dataset, and the hyperparameter chosen for the ensemble that was trained on all the features. 
We compared the performance of the ensemble trained on all the features to those trained when excluding each of the features.  

The results are presented in Table \ref{tab:FeatureContribution}. To assess the statistical significance, we used ANOVA models to examine the effect of excluding each feature on the accuracy, TPR and TNR. 
We found that excluding the velocity led to the largest difference in performance, resulting in significant decreases in the accuracy and TPR values, and an insignificant increase in the TNR (Table \ref{tab:statistics_feature}). Excluding the position led to significant increases in the accuracy and TNR values, and an insignificant decrease in the TPR. Omitting the gripper angle did not significantly change the performance relative to the full model. 

\begin{table}[h!]
  \begin{center}
    \caption{Feature Contribution}
    \label{tab:FeatureContribution}
    \resizebox{\columnwidth}{!}{
    \begin{tabular}{|c|c|c|c|c|}
      \hline
      \backslashbox{\textbf{Metric}}{\textbf{Excluded Feature}} & 
       \multicolumn{1}{|p{1.2cm}|}{\centering \textbf{No Excluded Feature}} &\multicolumn{1}{|p{1.2cm}|}{\centering \textbf{Position}} & \multicolumn{1}{|p{1.2cm}|}{\centering \textbf{Velocity}} & \multicolumn{1}{|p{1.2cm}|}{\centering \textbf{Gripper Angle}}\\
      \hline
      \textbf{Accuracy} & ${0.72} \pm {0.02}$ & ${0.76} \pm {0.01}$ & ${0.69} \pm {0.02}$ & ${0.72} \pm {0.01}$\\
      \hline
      \textbf{TPR} & ${0.74} \pm {0.02}$ & ${0.72} \pm {0.01}$ & ${0.66} \pm {0.03}$ & ${0.74} \pm {0.01}$\\
      \hline
      \textbf{TNR} & ${0.70} \pm {0.04}$ & ${0.80} \pm {0.02}$ & ${0.72} \pm {0.03}$ & ${0.69} \pm {0.01}$\\
      \hline
    \end{tabular}
    }
  \end{center}
\end{table}

\begin{table}[h!]
  \begin{center}
    \caption{Effect of Feature Exclusion on Performance}
    \label{tab:statistics_feature}
    \resizebox{\columnwidth}{!}{
    \begin{tabular}{|c|c|c|c|}
      \hline
      \backslashbox{\textbf{Metric}}{\textbf{Test}} & 
      \multicolumn{1}{|p{3cm}|}{\centering \textbf{Accuracy}} & \multicolumn{1}{|p{3cm}|}{\centering \textbf{TPR}} & \multicolumn{1}{|p{3cm}|}{\centering \textbf{TNR}}\\
      \hline
      \textbf{ANOVA model} & \thead{$F_{(3, 27)}=31.36$ \\ $p<10^{-4}$} & \thead{$F_{(3, 27)}=35.32$ \\ $p<10^{-4}$} & \thead{$F_{(3, 27)}=32.14$ \\ $p<10^{-4}$} \\
      \hline
      \textbf{Full vs. No Position} & \thead{$t_{(27)}=5.87$ \\ $p_{corrected}<10^{-4}$} & \thead{$t_{(27)}=1.96$ \\ $p_{corrected}=0.121$} & \thead{$t_{(27)}=8.29$ \\ $p_{corrected}<10^{-4}$}\\
      \hline
      \textbf{Full vs. No Velocity} & \thead{$t_{(27)}=3.73$ \\ $p_{corrected}=0.002$} & \thead{$t_{(27)}=8.43$ \\ $p_{corrected}<10^{-4}$} & \thead{$t_{(27)}=2.17$ \\ $p_{corrected}=0.078$}\\
      \hline
       \textbf{Full vs. No Gripper} & \thead{$t_{(27)}=0.24$ \\ $p_{corrected}=0.810$} & \thead{$t_{(27)}=0.86$ \\ $p_{corrected}=0.396$} & \thead{$t_{(27)}=0.38$ \\ $p_{corrected}=0.705$}\\
      \hline
      \textbf{No Position vs. No Velocity} & \thead{$t_{(27)}=9.60$ \\ $p_{corrected}<10^{-4}$} & \thead{$t_{(27)}=6.47$ \\ $p_{corrected}<10^{-4}$} & \thead{$t_{(27)}=6.12$ \\ $p_{corrected}<10^{-4}$}\\
      \hline
       \textbf{No Position vs. No Gripper} & \thead{$t_{(27)}=5.63$ \\ $p_{corrected}<10^{-4}$} &  \thead{$t_{(27)}=2.82$ \\ $p_{corrected}=0.027$} & \thead{$t_{(27)}=8.67$ \\ $p_{corrected}<10^{-4}$}\\
      \hline
      \textbf{No Velocity vs. No Gripper} & \thead{$t_{(27)}=3.97$ \\ $p_{corrected}=0.001$} &  \thead{$t_{(27)}=9.29$ \\ $p_{corrected}<10^{-4}$} &  \thead{$t_{(27)}=2.55$ \\ $p_{corrected}=0.050$}\\
      \hline
    \end{tabular}
    }
  \end{center}
\end{table}
\section{DISCUSSION}
In this work, we developed an ensemble of neural networks to predict when camera movements will occur in RAMIS using the kinematic data of the three PSMs. We did this by splitting the data into segments, and labeling each either as a segment that immediately precedes a camera movement, or one that does not. Due to the large imbalance between the number of \textit{before camera movement segments} and \textit{not before camera movement segments}, we used an ensemble of networks. We showed that the kinematic data can indeed be used to predict camera movements, and found which segment duration and ensemble size were best for the task with our dataset. Additionally, we found that advance prediction of camera movements in RAMIS can be possible.

The accuracy we achieved on the test set was 0.72, which does indeed show that the kinematic data is indicative of camera movements, as chance level in this case is 0.50. However, these results leave room for improvement. There are several potential explanations for this performance level. One is the fact that our dataset was very small; more data would likely improve the ensemble's performance. Available and labeled surgical data is a common bottleneck \cite{ross2018exploiting, van2020multi, wang2018satr}, and therefore poses a challenge when using deep learning to analyze surgical data. However, this work is the first to use RAMIS data to predict the timing of the camera movements, and demonstrates that it is possible to do so using the kinematic data of the surgical instruments. As the availability of data increases, the potential for better performance increases. Therefore, this work serves as a proof of concept, and the performance would need to be improved to be used in RAMIS.

An additional fact that may have contributed to the performance level is that this dataset was comprised of seven surgeons at two expertise levels (novices and experts), and
the camera movements likely differ between the surgical levels. For example, experts have been shown to make smaller and more frequent camera movements than novices \cite{jarc2017viewpoint}. We observed a similar tendency in our dataset, in which the experts made on average 35 and 57 camera movements in the “Uterine Horn Dissection” and “Simulated Cuff Closure” tasks, respectively, whereas novice surgeons made on average 10 and 24 camera movements in the two tasks. However, due to the small size of our dataset we could not explore the potential benefit of tailoring the network to the expertise of the user or other user specific factors such as school. This could be an interesting direction for future research, and we believe that training networks separately for different surgical expertise levels may improve the performance of the ensemble.

Moreover, this dataset contained two different procedures. More data would additionally allow for training networks separately on different procedures, which may be characterized by different movements. We can additionally raise the question of if these surgical procedures are representative of other procedures. That is, would this approach work for other surgical procedures? 
In this work, we hypothesized that predicting the timing of the camera movements from the instruments' kinematics might be possible.
Despite the small size of our dataset, we found this to be possible. We do not believe that neural networks trained on two specific procedures are representative of all the surgical procedures, and believe that there would be great value in training networks for specific procedures. However, as a relation between surgical instruments and camera movements likely exists in the majority of surgical procedures, we hypothesize that this method can be used for different surgical procedures. We believe that this is a hypothesis that is important to assess in future works.  

The fact that the kinematics of the instruments can be indicative of camera movements is consistent with previous works that designed systems in which the camera followed the instruments \cite{mariani_experimental_2020, eslamian_autonomous_nodate, eslamian2016towards, wijsman_first_2018}. In these works, they found that the autonomous systems were preferred by users and improved users' performance in dry-lab tasks.
However, similar to these works, ours too has the drawback of assuming the instruments to be the only deciding factor in RAMIS camera movements, while in reality, other features, such as anatomical features, likely affect the desired viewpoints \cite{rivas2019transferring, ko2005intelligent}. Therefore, in our future work, we may find it beneficial to add the visual information to the network, such that it can take the entire surgical scene into account. 
However, unlike systems in which the camera tracks the surgical instruments, our predictive method does not assume a specific method to be optimal. That is, having the camera follow the instruments assumes that the camera control method should be to follow the instruments. 
However, this may not be the method of camera control naturally employed by surgeons, or the method that would be most beneficial for them. Several works discuss the lack of flexibility of these systems \cite{rivas2019transferring, ko2005intelligent}. Training neural networks to predict when the surgeon would move the camera assumes the surgeon to be the model to imitate. That is, there is no assumption that the tools should be followed in a specific manner. Rather, the goal is to move the camera at the times the surgeon would have chosen, such that they achieve new viewpoints at the times they would have chosen, potentially allowing them to focus more fully on the surgical task.

In this work, we found the use of an ensemble to be an appropriate solution for the class imbalance. This is in accordance with \cite{zhang2015ensemble} and \cite{wang2015large}, which both used an ensemble of models with unbalanced datasets. Furthermore, by comparing between the performance of the ensemble and each of its individual composing networks, we found that the ensemble led to better and more consistent results than using only under-sampling to deal with the class imbalance (i.e., training only a single network on a sub-set of the data). We will note that we explored the possibility of under-sampling the majority class to different ratios relative to the minority class, and accounting for the imbalance in the loss calculation, however found that this led to lower TPR values than the ensemble.

Commonly, ensembles are comprised of different models \cite{zhang2015ensemble, wang2015large}. This can be beneficial, as different models may perform better in different aspects of the problem \cite{perrone1992networks}. In our work, the ensemble was comprised of networks with the same architecture, similar to \cite{cheng2020random}. We chose not to use different networks in our ensemble due to the very small size of our dataset. We posited that examining many possible network combinations and choosing the one with the highest performance on the \textit{validation} set might not lead to the best performance on the \textit{test} set, rather might be tailored to the small \textit{validation} set.

After creating ensembles comprised of 15 networks, we examined four segment durations to find which would achieve the best results. The segment durations we tested were 0.5, 1, 2, and 4 seconds. We found that the 1 s segment duration yielded the best results. There are several potential explanations for this result. First, the dataset used for the 1 s segment duration was larger than those of the 2 s and 4 s segment durations. It is therefore a possibility that this contributed to the superior performance of the 1 s segment dataset. On the other hand, the 0.5 s duration dataset was larger than the 1 s duration dataset, and yet performed less well in terms of accuracy and TNR. We hypothesize that the 0.5 s duration may not have contained enough information to differentiate between \textit{before camera movement} and \textit{not before camera movement} segments. 

An additional possible reason that the longer duration sets performed less well can be that the \textit{before camera movement} and \textit{not before camera movement} segments may also have been similar in the cases of the 2 s and 4 s segments. That is, these segments may have contained information indicative of camera movements. However, this information may have been only part of the segment, while the rest of the segment could have been unrelated to a potential upcoming camera movement. If this were to be the case, the added information, which was unrelated to the camera movements, could have made it harder for the network to differentiate between \textit{before camera movement} and \textit{not before camera movement} segments, leading to the observed decrease in performance. 

Yet another potential explanation is the overlap that could exist between segments. We segmented the dataset with an overlap of N-1. Following that, we trained each network on a balanced subset of the data by using all the \textit{before camera movement segments}, and a randomly selected subset of \textit{not before camera movement segments}. There were many more \textit{not before camera movement segments}, such that due to the random selection of our subsets, many of the segments did not overlap with each other. However, the longer segment duration datasets did have more overlapping segments than the shorter duration datasets.
Hence, this may too have contributed to the superior performance of the shorter durations. Future work with larger datasets can allow for better isolation of the causes that led to the better performance of the 1 s segment dataset. Additionally, larger datasets, as well as datasets containing additional surgical procedures, can allow for testing if there are segment durations that consistently lead to the best results. 

The 1 s segment duration may appear short. However, works that have decomposed surgical tasks (e.g., suturing) into their composing gestures have shown that there are gestures which fall in the range between one and two seconds, and some are even slightly shorter than one second \cite{itzkovich2019using, dipietro2016recognizing}. This indicates that meaningful surgical data can be included in data segments of one second. This is further supported by the fact that reaction times in surgery have been shown to be under a half a second \cite{drag2010cognitive, zheng2003reaction}, and therefore information indicating an upcoming camera movement can be included in a one-second segment. These studies showed that the reaction times were generally above 300 msec, further supporting the idea that the 0.5 s segments likely did not contain enough information to be indicative of camera movements.

We examined the accuracy of our ensemble, however it was also important to separately assess the ability of the ensemble to recognize an upcoming camera movement when there was one (as quantified by the TPR), as we are interested in a system that may be able to initiate a camera movement at the appropriate time. However, of equal importance is the ability of the ensemble to predict no imminent camera movement when there is none (as quantified by the TNR). If the system were to move the camera at an inappropriate time, it could disorient and confuse the surgeon, as well as cause critical features or instruments to be outside the field of view. 
The ensemble achieved a TPR of 0.74 and a TNR of 0.70 on the \textit{test} set. Similar to the accuracy, these metrics need to be further improved to be used in RAMIS, however are above chance level, and show that the ensemble learned how to recognize both \textit{before camera movement} and \textit{not before camera movement} segments, as desired. 

We also were interested in attempting to understand the contribution of each of the features to the performance of the network. 
We found that removing the velocity led to decreases in the accuracy and TPR values, whereas excluding the position led to increases in the accuracy and TNR values. These results indicate that the velocity might be important for predicting the timing of camera movements in RAMIS. If the velocity is the most indicative feature, removing the position signals might improve the performance of the ensemble. This decreases the number of features, hence decreasing the size of the network, and potentially leaving only the most relevant information present. Omitting the gripper angle led to similar performance to that of the full network, indicating that this feature may not provide additional information. Furthermore, these results indicate the robustness of the ability of the ensemble to learn the timing of the camera movements using the kinematics of the instruments in our dataset. However, further work with larger datasets is needed to fully ascertain which of the features are consistently most indicative of the camera movements, and what relations between the features are learned by the neural networks. 

The end goal of our work is to develop an autonomous camera controller. We trained an ensemble to recognize if a camera movement will, or will not,  follow a segment of kinematic data. Hence, no data from after the initiation of the camera movement was used by our ensemble. This is in contrast to other studies that inputted both past and future RAMIS data points into neural networks in classification problems \cite{itzkovich2019using, dipietro2016recognizing}. Therefore, our proposed ensemble can potentially be used for online prediction of upcoming camera movements by feeding segments of kinematic data into the trained ensemble during RAMIS, and recognizing online which segments should be followed by a camera movement. 

Nevertheless, when designing such an autonomous controller, instructing the camera to move immediately would not be feasible. Therefore, we examined the possibility of predicting that a camera movement will happen in a certain amount of time rather than in the next sample. We found that advance prediction of camera movements is possible. Predicting a camera movement 0.25 s in advance led to only a slight decrease in performance, such that all three metrics were at least 92\% relative to imminent prediction. Specifically, we observed no significant decrease in accuracy, whereas there was a decrease in the TPR and an increase in the TNR relative to imminent prediction. Predicting camera movements 0.5 s in advance led to a slightly larger decrease in the performance, however, here too, all three metrics were close to, or higher than, 90\% relative to immediate prediction. Despite this, the decreases in accuracy and TPR were statistically significant. Predicting camera movements 1 s in advance led to a larger decrease in the performance, such that the accuracy of predicting a camera movement 1 s in advance was 84\% relative to imminent prediction, i.e., an accuracy level of 0.55. Based on these results, we posit that predicting a camera movement 1 s in advance may lead to too large of a decrease in the reliability of the prediction. However, we also note that a larger dataset would allow for more precise characterization of the limit of advance prediction. Even with the small dataset, our results demonstrate that camera movements can be predicted slightly in advance, with only a small decrease in performance, supporting the possibility of one day performing online camera movement prediction in RAMIS. 

In addition to the requirements of improving the performance and using larger datasets \cite{davenport2019potential}, prior to potentially using such an autonomous system in RAMIS, there are additional challenges and limitations when considering adopting AI (artificial intelligence) based systems in the medical field. For example, it is vital to ensure the safety of such a system by conducting surgical training experiments in which surgeons compare manual camera control to autonomous camera control. It will also be important to hear if surgeons prefer the autonomous system. It can additionally be important to study potential affects that the optics of the camera may have on camera movements and their prediction. 


There are also ethical questions that arise when considering the use of AI in the medical field. For example, there is the question of interpretability. While a surgeons may be able to explain the rational for their decisions, explaining the cause of an autonomous system's output can be more challenging \cite{davenport2019potential, o2020bias}. There is additionally the question of accountability in the event of a mistake made by an AI system \cite{davenport2019potential}. Furthermore, there is the risk of bias, in which the system may have biases in their predictions relating to features that are not actually causal factors in that case, such as gender or race \cite{o2020bias, safdar2020ethical}. 
There is also the question of rare cases; while the surgeon may be able to use their knowledge and experience to adapt to a rare or unexpected case, an AI system may not know how to respond to an unfamiliar case \cite{safdar2020ethical, shaw2019artificial}. Hence, there are many ethical issues to consider when developing AI systems for medical use.

However, AI has also been shown to be beneficial in medical use. For example, AI has been shown to improve patient outcomes, and has the potential to reduce the workload of medical personnel \cite{shaheen2021ai}. We believe that a system which can autonomously move the camera has the potential to benefit surgeons and to reduce their workload. However, such a system would need to be trained on a very large and varied dataset \cite{itzkovich2021generalization}, as well as tested extensively. Furthermore, we believe that such a system would still need to allow control for the surgeon. For example, the surgeon should be able to move the camera on their own should they feel it appropriate. Hence, the potential introduction of such a system into surgeries has many preceding steps. However, this work shows that it may one day be possible, as well as sheds scientific light of the possibility of predicting the timing of camera movements made by surgeons using the instruments' kinematics. 

\section{CONCLUSIONS}
In this work we used an ensemble of LSTM neural networks to analyze recordings of surgical tasks performed with the da Vinci surgical system, and demonstrated that the kinematic data of the surgical instruments can be used to predict when camera movements will occur. We found the segment duration and ensemble size that produced the best results on our dataset and tuned the hyperparameters of the networks in the ensemble. Furthermore, we showed that, in addition to the prediction of imminent camera movements, the camera movements can be predicted slightly in advance. Our findings may be the first step towards designing a predictive autonomous camera controller, which would allow robotic surgeons to benefit from optimal viewpoints without necessitating the frequent camera movements they currently have to make manually. 
\addtolength{\textheight}{0cm}   

\section*{ACKNOWLEDGMENTS}
We would like to thank Intuitive Surgical Inc. for providing us with the RAMIS data, and Anthony Jarc and Yarden Sharon for their valuable insights.

\ifCLASSOPTIONcaptionsoff
  \newpage
\fi




\bibliographystyle{IEEEtran}
\bibliography{PhD}
%


%







\end{document}